\documentclass[lettersize,journal]{IEEEtran}
\usepackage{amsmath,amsfonts}
\usepackage{algorithmic}
\usepackage{algorithm}
\usepackage{array}
\usepackage[caption=false,font=normalsize,labelfont=sf,textfont=sf]{subfig}
\usepackage{textcomp}
\usepackage{stfloats}
\usepackage{url}
\usepackage{verbatim}
\usepackage{cite}
\usepackage{multirow}
\usepackage{colortbl}
\usepackage{graphicx}
\usepackage{booktabs} 
\usepackage{verbatim}
\usepackage[table]{xcolor}
\newcommand{\etal}{\emph{et al.}}
\usepackage{makecell}
 
\usepackage[switch]{lineno}

\usepackage{hyperref}
\hypersetup{
    colorlinks=true,
    citecolor=blue,
    linkcolor=blue,
    urlcolor=blue
}

\begin{document}

\title{Cross-Domain Knowledge Distillation for Low-Resolution Human Pose Estimation}

\author{Zejun Gu, Zhong-Qiu Zhao, Henghui Ding, Hao Shen, Zhao Zhang, \IEEEmembership{Senior Member, IEEE}, and De-Shuang Huang, \IEEEmembership{Fellow, IEEE}
\thanks{Zejun Gu, Zhong-Qiu Zhao, Hao Shen, and Zhao Zhang are with the School of Computer Science and Information Engineering, Hefei University of Technology, Hefei 230009, China. (e-mail: guzejunmail@gmail.com, haoshenhs@gmail.com; cszzhang@gmail.com; Corresponding author: Zhong-Qiu Zhao)}
\thanks{Henghui Ding is with the School of Computer Science, Fudan University, Shanghai 200433, China (e-mail: henghui.ding@gmail.com)}
\thanks{De-Shuang Huang is with the Institute of Machine Learning and Systems Biology, Eastern Institute of Technology (e-mail: huangdeshuang@163.com)}
}

\markboth{Journal of \LaTeX\ Class Files,~Vol.~14, No.~8, August~2021}%
{Shell \MakeLowercase{\textit{et al.}}: A Sample Article Using IEEEtran.cls for IEEE Journals}


\maketitle

\begin{abstract}
In practical applications of human pose estimation, low-resolution inputs frequently occur, and existing state-of-the-art models perform poorly with low-resolution images. This work focuses on boosting the performance of low-resolution models by distilling
knowledge from a high-resolution model. However, we face the challenge of feature size mismatch and class number mismatch when applying knowledge distillation
to networks with different input resolutions.
To address this issue, we propose a novel cross-domain knowledge distillation (CDKD) framework. In this framework, we construct a scale-adaptive projector ensemble (SAPE) module to spatially align feature maps between models of varying input resolutions. It adopts a projector ensemble to map low-resolution features into multiple common spaces and adaptively merges them based on multi-scale information to match high-resolution features. Additionally, we construct a cross-class alignment (CCA) module to solve the problem of the mismatch of class numbers. By combining an easy-to-hard
training (ETHT) strategy, the CCA module further enhances the distillation performance. The effectiveness and efficiency of
our approach are demonstrated by extensive experiments on
two common benchmark datasets: MPII and COCO.
 The code is made available in supplementary material.
\end{abstract}

\begin{IEEEkeywords}
Knowledge distillation, low-resolution image, human pose estimation, cross-domain distillation.
\end{IEEEkeywords}

\section{Introduction}
\IEEEPARstart{H}{uman} pose estimation (HPE) is a fundamental task in computer vision which aims to predict the positions of body joints from RGB images~\cite{geng2023human,ye2023distilpose,qu2022heatmap,xu2022vitpose,yang2021transpose,yang2016end}. The recent progress has focused on training methods~\cite{iqbal2020weakly,schmidtke2021unsupervised}, network structures~\cite{sun2019deep,wang2020deep,li2021tokenpose}, 
and fusion strategies~\cite{cheng2022dual}, which have notably advanced the accuracy of HPE with high-resolution images~\cite{liu2022ehpe,li2022exploiting,li2019multi,zou2022human,ning2017knowledge}.

In real-world application scenarios, images are usually captured in low resolutions, for example, wide-view video surveillance
and long-distance shooting. In addition, high-resolution input will bring great
computational and memory complexity, which impedes the development of practical applications. 
However, when current state-of-the-art models are directly applied to low-resolution images, significant performance degradation occurs due to the lack of rich image information. Therefore, it is a critical yet more challenging problem to upgrade the performance of a low-resolution HPE model.
One possible way to solve the problem is to transfer the knowledge from high-res models to low-res models. 
\begin{figure}[!t]
\centering	\includegraphics[width=0.99\linewidth,height=0.95\linewidth]{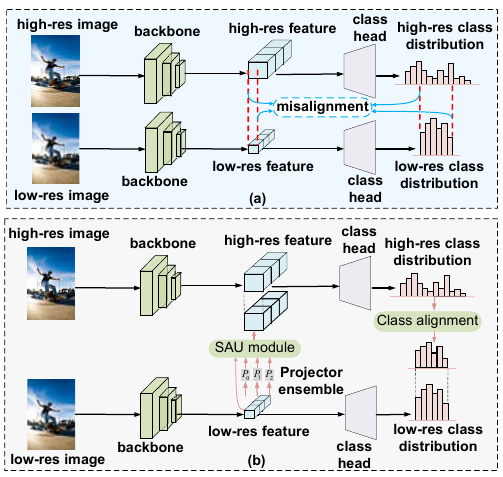}
\caption{
Comparison between traditional distillation and our CDKD. 
 \textbf{(a)}  Illustration of the issue of traditional methods for distillation between different resolution models. When distilling knowledge from high-res models to low-res models, there is a problem of misalignment between features and output classes. \textbf{(b)} 
 The overview of our proposed CDKD framework. }
\label{fig01}
\end{figure}


However, the different input resolutions of the teacher model and the student model lead to the following two problems: 1) The teacher model and the student model do not share the same feature spatial size at the same network stages. In this situation, traditional feature distillation methods can not work. 2) The number of output categories of the teacher model and the student model are not the same. Currently, there are \textit{\textbf{NO}} logit distillation studies based on different class numbers in detection tasks, so the logit information in teacher models can not be effectively distilled. 


Due to the mismatch in the feature spatial domain and output distribution domain between the teacher model and the student model, we propose a novel knowledge distillation framework termed cross-domain knowledge distillation (CDKD) to resolve it. It mainly includes
the following two components:
1) A scale-adaptive projector ensemble (SAPE) module for feature distillation. 2) A cross-class alignment (CCA) module for logit distillation.

\begin{figure}[htbp]
\centering	\includegraphics[width=1.0\linewidth,height=0.701\linewidth]{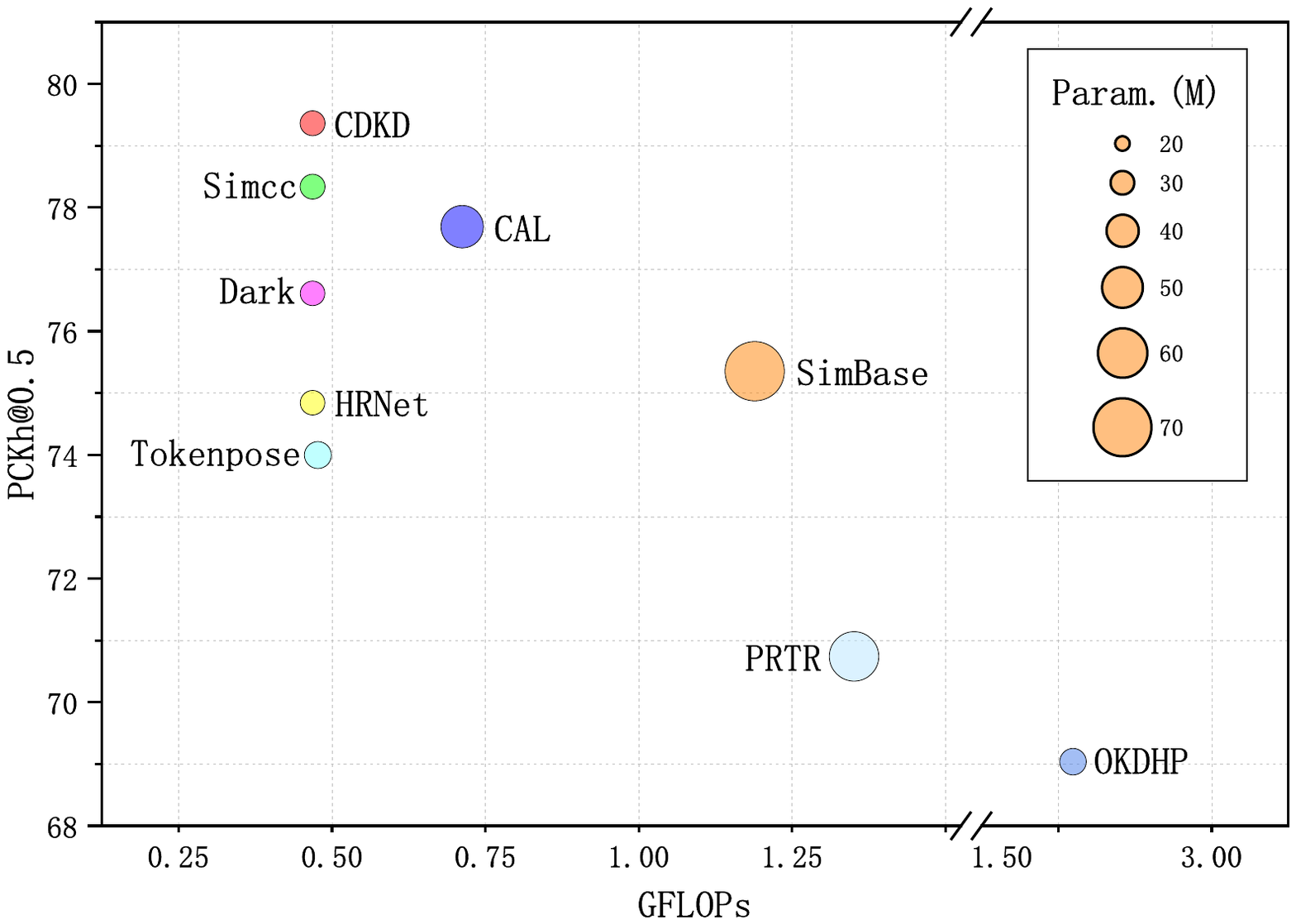}
\caption{Comparisons between the SOTA methods and the
proposed CDKD on the MPII~\cite{andriluka20142d} val dataset with the input resolution of 64$\times$64.
 Red circles at the
upper left corner denote CDKD. 
It outperforms SOTA
models in terms of accuracy (AP), Parameter, and computational cost (GFLOPs).
}
\label{figksh}
\end{figure}

\IEEEpubidadjcol
The SAPE module resolves the feature size mismatch between the high-res teacher
and the low-res student.
Firstly, we design a projector to map features into a common space for matching. Feature distillation can be regarded as a multi-task learning process, including feature learning for the original task and feature matching
for distillation~\cite{chen2022improved}. In this scenario, the student network might overfit the teacher's feature distributions, leading to less distinguishable generated features for HPE.
Adding a projector for distillation would alleviate the overfitting issue. We propose to ensemble projectors for further improvement.
According to the theory about ensemble learning~\cite{zhou2002ensembling}, different initialized projectors produce different transformed features.
The use of multiple projectors helps to improve the generalization ability of the student. To address the
body scale variation problem in the natural scene, we propose a scale-adaptive unit (SAU) 
to capture the multi-scale information. It assigns different weights to each projector by multiple parallel transformations with different receptive fields.
Then, it merges the output features of each projector with assigned weights.

The CCA module 
 solves the problem of class number mismatch. As shown in Fig.~\ref{fig01} (b), the probability distribution of the classification output for different resolution models has the same mathematical significance.
  It can be shared directly between the teacher model and the student model. Meanwhile, the probability values of multiple adjacent classes can be merged by adding them together. In addition, the probability distribution of the classification output is also a fundamental element in calculating distillation loss.
 Based on this observation, for the high-res teacher, we merge the probability values of adjacent categories to make them consistent with the categories of the student model. It allows for the implementation of logit distillation. 
 Finally, to enhance the effectiveness of distillation, we further propose an easy-to-hard training (ETHT) strategy.
It improves distillation performance through 
curriculum learning.

In summary, our CDKD achieves optimal performance with minimal computational cost. 
As shown in Fig.~\ref{figksh}, CDKD outperforms previous SOTA
methods, such as Simcc~\cite{li2022simcc} and CAL~\cite{wang2022low}
with few parameters and GFLOPs.

 Our contributions are summarized as follows:
 \begin{itemize}
 \item We propose a novel scale-adaptive projector ensemble (SAPE) module that solves the problem of feature size mismatch between the high-res teacher and the low-res student and attains excellent distillation performance.
  \item We construct a cross-class alignment (CCA) module to tackle the problem of inconsistent class numbers in logit distillation. Combined with the easy-to-hard training (ETHT) strategy, it further improves the distillation effectiveness.
  \item Our proposed CDKD is a universal framework that can address the issue of mismatched feature spatial domain and output distribution domain in different distillation tasks.  
  Extensive experiment results demonstrate its efficiency, effectiveness, and universality. It achieves SOTA performance in low-res HPE on MPII~\cite{andriluka20142d} and COCO~\cite{lin2014microsoft} with \textit{\textbf{NO}} computational cost increment.  
When applied to different HPE models and various low-res inputs, it consistently achieves superior performance. 
 \end{itemize}

\section{Related Work}
\subsection{Human Pose Estimation}

The current research on human pose estimation primarily focuses on two aspects: keypoint coordinate representation and applications under complex conditions.


For keypoint coordinate representation, recent works are mainly divided into three mainstreams: regression-based methods~\cite{li2021human,li2021pose, belagiannis2015robust,carreira2016human,tian2019directpose,toshev2014deeppose}, heatmap-based methods~\cite{li2019rethinking,wang2022lite,xiao2018simple,bulat2016human,tompson2014joint,zhang2020distribution,li2021tokenpose},  and new keypoint representation methods~\cite{li2022simcc,geng2023human}. 
Regression-based methods directly regress the coordinates of keypoints within a lightweight framework.
Deeppose~\cite{toshev2014deeppose} is the first to propose direct regression of joint coordinates.
 CenterNet~\cite{zhou2019objects} 
 presents a method to accomplish multi-person pose estimation within a one-stage object detection framework, directly regressing joint coordinates instead of bounding boxes. 
SPM~\cite{nie2019single} introduces root joints for distinguishing among different person instances, by employing hierarchical rooted representations of human body joints to improve the prediction of long-range displacements for specific joints.
The residual log-likelihood (RLE)~\cite{li2021human} utilizes normalizing flows to capture the underlying output distribution, which enables regression-based methods to achieve accuracy comparable to SOTA heatmap-based methods.
MDN~\cite{varamesh2020mixture} proposes a mixture density network for regression.
Regression-based methods have significant advantages in speed, but their accuracy is insufficient. 

The heatmap-based methods adopt a two-dimensional Gaussian distribution to represent joint coordinates.
Some studies optimize backbones to extract better features.
Sun \etal~\cite{sun2019deep} introduce a groundbreaking network designed to preserve high-resolution representations throughout the entirety of the process, resulting in substantial performance enhancements.
Other studies aim to improve prediction accuracy by reducing quantization errors.
Zhang \etal~\cite{wang2022low} propose using a distribution-aware coordinate representation for post-processing of prediction results to reduce quantization errors.
Huang \etal~\cite{huang2020devil} design a plug-and-play unbiased data processing method that effectively enhances the performance of different models without increasing computational complexity.
Some methods~\cite{li2021tokenpose,xu2022vitpose} improve their performance by leveraging transformers because they can capture long-range information.
Heatmap-based methods are far ahead in terms of performance, but they have the disadvantage of exceptionally high computational cost and slow preprocessing operations. 
DistilPose~\cite{ye2023distilpose} combines heatmap and regression methods through knowledge distillation, achieving high accuracy at low computational costs.

New keypoint representation methods
explore novel approaches of keypoint representation to reduce quantization errors and leverage keypoint relationships. 
PCT~\cite{geng2023human} represents a pose by discrete tokens to model the dependency between the body joints.
Simcc~\cite{li2022simcc} effectively minimizes quantization error by transforming the keypoint regression task into a classification problem.
They all open up new perspectives on keypoint representation methods.

In addition, many studies are beginning to focus on applications under complex conditions. 
Yang \etal~\cite{yang2023explicit} introduce ED-Pose, which achieves fast, concise, and accurate end-to-end human pose estimation.
 Lee \etal~\cite{lee2023human} present ExLPose to tackle the problem of current models failing to work properly under low-light conditions.
Ju \etal~\cite{ju2023human} 
propose a dataset for human pose estimation in artwork, aiming to apply the human pose estimation model to virtual scenes.
 Yang \etal~\cite{yang2023effective} introduce an efficient full-body pose estimation method (DWPose), aiming at better application in human-computer interaction.
Sun \etal~\cite{sun2024aios} propose the first all-in-one-stage model for
expressive human pose and shape estimation, which achieves the best performance.


\subsection{Low-Resolution Vision Tasks}
Wang \etal~\cite{wang2022low} propose a novel confidence-aware learning (CAL) method to reduce quantization errors, thereby improving the performance of human pose estimation models under low-resolution conditions.
Li \etal~\cite{li2022simcc} design a simple coordinate classification 
method, which achieves excellent results in low-resolution pose estimation by transforming coordinate regression into a classification problem.
Kumar \etal~\cite{kumar2020s2ld} present a semi-supervised approach to predict landmarks on low-resolution images by
learning them from labeled high-resolution images. It can improve performance on the critical task of face verification in low-resolution images.
Chai \etal~\cite{chai2023recognizability} introduce a recognizability embedding enhancement approach to address the very low-resolution face recognition (VLRFR) challenge.
Sunkara \etal~\cite{sunkara2022no} design a new CNN building block (SPD) for low-resolution
images and small objects. This block is universal and can replace the downsampling layers in different CNN models, thereby improving the performance of various low-resolution tasks. 
Xu \etal~\cite{xu20203d} propose a resolution-aware neural network for human pose estimation which can deal with different resolution images with a single model.

\subsection{ Knowledge Distillation}
Knowledge distillation aims to transfer knowledge from a trained teacher model to a compact and lightweight student model. In recent years, various methods have been proposed for knowledge distillation~\cite{zhu2023scalekd,yang2023effective,ni2023dual}. These methods fall into two lines of work: 1) feature distillation methods~\cite{huang2023generic}. 2)
logit distillation methods~\cite{zhao2022decoupled}. Feature distillation methods distill knowledge using intermediate features, while logit distillation methods are designed to perform distillation on output logits.

Logit distillation
was originally proposed using the KL divergence~\cite{hinton2015distilling}, and
it has been extended using spherical normalization~\cite{guo2020reducing}, label decoupling~\cite{zhao2022decoupled}, and probability reweighting~\cite{niu2022respecting}.
MLD~\cite{jin2023multi} introduces a multi-level prediction alignment framework to logit distillation.
Through this framework,
the student model learns instance prediction, input correlation, and category correlation simultaneously.
LD~\cite{zheng2022localization} presents a novel localization distillation (LD) method that can efficiently transfer the localization knowledge from the teacher to the student.

While logit distillation methods may seem straightforward and versatile for application across various scenarios, their performance frequently falls short compared to feature distillation.
Compared to logit distillation, feature distillation methods are more likely to achieve high performance because they absorb rich knowledge from the teacher model.
The work~\cite{miles2024v_kd} adopts a novel orthogonal projection layer to maximize the distilled knowledge to the student backbone, thereby achieving outstanding performance.
CrossKD~\cite{wangcrosskd} transfers the intermediate features from the student's head to that of the teacher, generating crosshead predictions for distillation.
This approach efficiently mitigates conflicts between supervised and distillation targets.
Some methods~\cite{heo2019comprehensive,heo2019knowledge}  mitigate the differences in features between the teacher and student models, thus compelling the student model to emulate the teacher model at the feature level. Other methods~\cite{park2019relational,tian2019contrastive} transmit teacher knowledge by extracting input correlations.

In contrast to existing KD methods that
focus on improving the performance of lightweight networks, this paper aims to improve the low-res
model by distilling from the high-res model. 
Recently, a succession of relevant studies has emerged.
The work~\cite{qi2021multi} adopts a multi-scale aligned distillation method to tackle the output size mismatch, but this method relies on the feature pyramid (FPN) structure~\cite{lin2017feature} as its backbone. The distillation method improves the low-res model by using upsampling in the initial stage~\cite{shin2022teaching}. However, upsampling introduces noise and incurs substantial computational costs. FMD~\cite{huang2022feature} solves the problem of inconsistent feature sizes, but it is under the limited condition that the number of channels will not remain constant.
FITNETS~\cite{romero2014fitnets} utilizes convolutional downsampling to align high-res features with their low-res counterparts.  Nonetheless,
downsampling drastically reduces the rich efficient information hidden in large feature maps.
ScaleKD~\cite{zhu2023scalekd} designs a mapping function to
align the size of the feature map in different models, but it does not achieve outstanding performance. 
Our proposed method (CDKD) can solve the above problems and thereby achieve more competitive distillation effects.

\section{Method}
\begin{figure*}[htbp]
	\centering	\includegraphics[width=0.99\linewidth,height=0.48\linewidth]{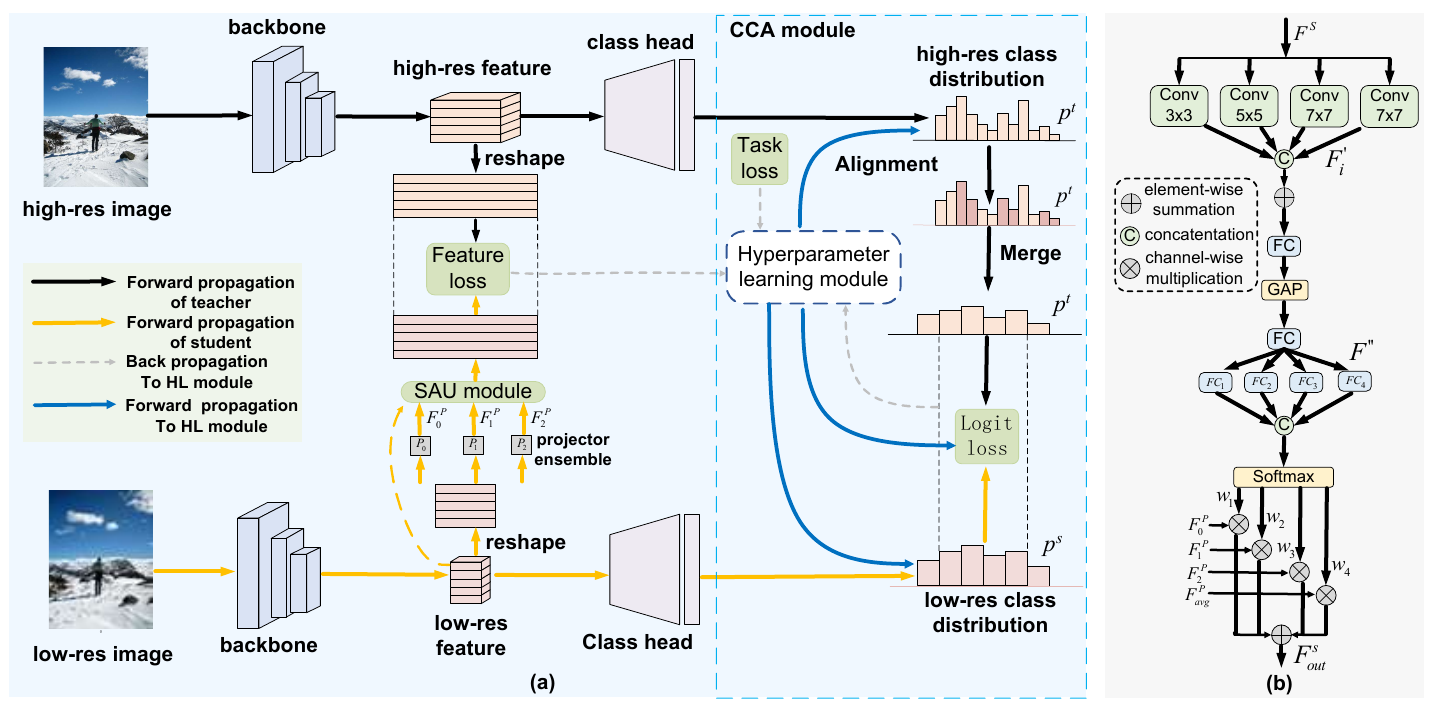}
	\caption{\textbf{(a)} Overall architecture of our proposed CDRD. 
 During training, a well-trained and fixed high-res teacher provides rich knowledge to
help the training of a low-res student based on the alignment of features and classes.  \textbf{(b)} The illustration of our SAU module.
 }
	\label{fig02}
\end{figure*}

In this section, we propose a distillation-based human
pose estimation framework, cross-domain knowledge distillation (CDKD), as shown in Fig.~\ref{fig02} (a). In CDKD, the teacher is a high-res model, and the student is a low-res model. We transfer the rich details and texture 
knowledge from the teacher model to the student model during training.
\subsection{Scale-Adaptive Projector Ensemble }
 The Scale-Adaptive Projector Ensemble (SAPE) aims to align the feature size of the teacher and the student, which consists of a projector ensemble and a scale-adaptive unit.

 \subsubsection{Projector Ensemble }
 As shown in Fig.~\ref{fig02} (a), our framework is based on the Simcc~\cite{li2022simcc} algorithm. Recent research indicates that feature distillation is more likely to achieve superior performance compared to logit distillation~\cite{jin2023multi}. The last feature of
networks is better suited for distillation~\cite{deng2021comprehensive}. One possible reason is that the last feature is closer to the classifier and will directly impact classification effectiveness. Therefore, we adopt the last feature to distill.

We represent the last teacher feature 
as $F^{t}\in\mathbb{R}^{B\times C\times H^{\prime}\times W^{\prime}}$, where
$\mathit{B}$, $\mathit{C}$, $\mathit{H^{\prime}}$, and $\mathit{W^{\prime}}$ are the batch size, the number of channels, the height, and the width of the last teacher feature maps, respectively. 
The corresponding student feature is represented by
$F^{s}\in\mathbb{R}^{B\times C\times H\times W}$, where
$\mathit{B}$, $\mathit{C}$, $\mathit{H}$, and $\mathit{W}$ are the batch size, the number of channels, the height, and the width of the last student feature maps, respectively. 
We define $\mathit{m}$ as the scale factor between the high and low resolution, where $\begin{aligned}H=H^{\prime}/{m}\end{aligned}$ and $\begin{aligned}W=W^{\prime}/{m}\end{aligned}$.

In classification tasks based on distillation, the training process
of the student network can be considered as multi-task learning within the same feature space. Therefore, student features are prone to overfit teacher
features and would be less discriminative for classification. 
We add a projector $P_i$ to disentangle the two tasks to improve the student's performance and to match the size of  $F^{t}$ (see Fig.~\ref{fig02} (a)).
In addition, we use an ensemble of projectors for further improvement which consists of a group of parallel projectors. Each projector contains a FC layer and a ReLU function. 

There are two
motivations for adopting ensemble projectors. 1) Projectors with diverse initializations yield distinct transformed features, contributing to the generalizability of the student. 2) The projected student features may include zeros due to the use of the ReLU function in the projector. In contrast, teacher features are
less likely to be zeros 
since the average pooling operation
is widely employed in CNNs. The feature distribution gap between the teacher model and the student model is large when using a single projector.
In this paper, we use ensemble learning to reduce the feature distribution gap and 
achieve better generalizability.

 \subsubsection{Scale-Adaptive Unit }
The previous work~\cite{chen2022improved} uses a simple additive approach
to fuse output features from different projectors. The simple additive approach not only fails to capture multi-scale information in real-world scenarios but also lacks the weight assignment based on the importance of different projectors. Inspired by~\cite{li2019selective}, we
propose the scale-adaptive unit (SAU) to deal with the problem.
The SAU learns to combine
all outputs from different projectors to produce a rich output feature.
 It is composed of multiple parallel transformations with different receptive fields, exploiting local and global information to obtain a strong low-res feature. 
 
 The architecture of the proposed SAU
is depicted in Fig.~\ref{fig02} (b).
We take the output of each projector and the low-res feature as input of SAU.
Initially,  we operate it through convolutions with different kernel sizes:

\begin{figure}[t]
	\centering	\includegraphics[width=1.00\linewidth,height=1.08\linewidth]{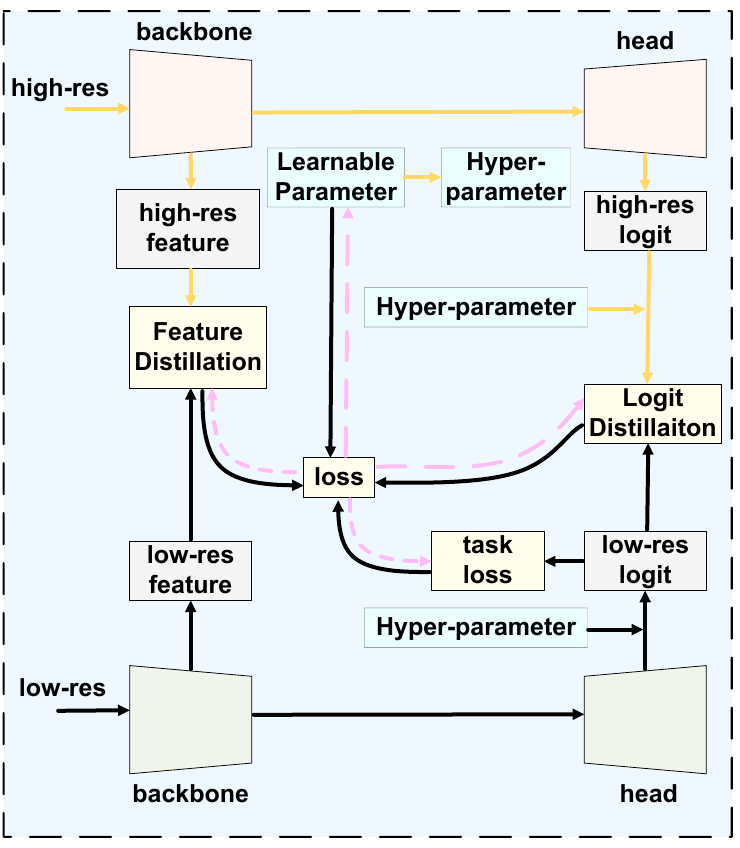}
 
	\caption{The illustration of our proposed easy-to-hard training (ETHT) strategy.}
	\label{fig03}
\end{figure}

\begin{equation}{F}_i^{^{\prime}}={Con}v_i({F}^s)\end{equation}
where  $\mathit{Conv}_i(\cdot)$ denotes convolution operation 
with different kernel sizes, $F^{s}$ is the last student feature, ${F}_i^{^{\prime}}$ is output of
each convolution operation and $\mathit{i}$ is the index of each branch.

Convolutions with different receptive fields help the model to obtain human body information at different scales. Because low-res images contain less useful information, the feature information under a small receptive field does not make sufficient sense. Therefore, we moderately increase the proportion of convolutions with a large receptive field to obtain as much useful information as possible, as shown in Table \ref{tab06}. 
Next, we fed ${F}_i^{^{\prime}}$ to fusion operation:
\begin{equation}F^{\prime\prime}=Fus(Concat[F_1^{^{\prime}},F_2^{^{\prime}},F_3^{^{\prime}},F_4^{^{\prime}}])\end{equation}
where $\mathit{Concat}[\cdot]$  and $\mathit{Fus}(\cdot)$  denote the concatenation operation and the fusion operation, respectively. This fusion operation is sequentially composed of a fully connected layer, a global pooling layer, and a summation layer. It can adaptively adjust receptive field sizes according to the image content. 
Then, $F^{\prime\prime}$ is applied to select operation:
\begin{equation}w_k=Select(F^{^{\prime\prime}})\end{equation}
where 
$\mathit{Select}(\cdot)$ is the select operation and ${w_k}$ is the weight value corresponding to each projector output. The select operation is sequentially composed of multiple fully connected layers, a concatenation layer, and a softmax layer. This operation can adaptively select different spatial scales of information with soft attention across channels. Finally, we fuse outputs from multiple projectors via an element-wise summation to obtain the
output feature $\mathit{F}_{out}^s$:
\begin{equation}\begin{aligned}{F}_{out}^s=\sum_{k=1}^K{w}_k\otimes{F^P_k}
\end{aligned}\end{equation}
where $\mathit{K}$ is the number of branches, $F^P_k$ is the output of each projector, ${w}_k$ is the corresponding weight
value, and  $\mathit{F}_{out}^s$ is the final output of this module.

After aligning features from multi-scale spaces, we can use cosine distance between $\mathit{F}_{out}^s$  and $\mathit{F}^{t}$ to calculate the feature distillation loss :

\begin{equation}\begin{aligned}
{L}_{fea}=
1-\frac{{F}_{out}^s}{\|{F}_{out}^s\|_2}\cdot\frac{{F}^{t}}{\|{F}^{t}\|_2}\end{aligned}\end{equation}
where $||\cdot||_2$ denotes
L2-norm.

\subsection{ Cross-Class Alignment }
Another way to transfer knowledge from teacher to student is logit distillation.
The feature distillation focuses on optimizing the encoder and does not directly impact the classification head, so we propose a cross-class alignment (CCA) module to refine the student's classification head straightforwardly.

\subsubsection{Class Alignment }
As is shown in Fig.~\ref{fig01} (a), the number of Simcc's output categories is proportional to the input resolution typically. The number of output categories of low-res model and high-res model is different when their resolution is different. However, the current logit distillation is based on the consistency of the number of categories. 
In our work, we propose a cross-class (CCA) alignment module to resolve this issue. 

Simcc's classification method can be regarded as one of ordinal regression~\cite{diaz2019soft}. There is a certain ordinal relationship among the categories in ordinal regression. Any number of adjacent classes can form an entire meaningful interval. We are interested in exploring the merging of multiple discrete categories of the student to align with the teacher. However, the values of the logit or the feature can not be added together, since they do not have the same computational unit and interpretable real-world meaning.

Combining the characteristics of ordinal regression and the properties of probability distributions, we have the following findings:
1) After the last softmax function, the values of the probability distribution can be added together.
2) The final output distribution of the model represents the probability of the keypoint at each relative position. Regardless of the resolution of images or the number of categories output by the model, the relative positions of keypoints in images are the same. 
3) The output probability distributions of images at different resolutions have the same units and practical significance. So they 
can be shared between different resolution models.
4) The probability distribution output by the model is also the subject of computation for traditional logit distillation loss which serves as the bridge connecting the teacher model and the student model.
Therefore, we adopt the class alignment method to align the category of different resolution models, as shown in Fig.~\ref{fig02} (a). Specially, we sum up the output probability values for every $\mathit{m}$ adjacent categories in the high-res teacher model:
\begin{equation}\boldsymbol{p}_j^t=\sum_{i=j\times m+1}^{j\times m+m}\boldsymbol{p}_i^t\end{equation}
where $\mathit{m}$ represents the ratio between high resolution and low resolution, $\boldsymbol{p}_i^t$ denotes the output probability values of i-th category in the high-res model,
\begin{table*}[htbp]
  \caption{Comparison with SOTA
  methods on the MPII val dataset.}
  \centering
    \renewcommand\arraystretch{1.25}
    \setlength{\tabcolsep}{3.0pt}
    \begin{tabular}{ccccccccccccc}
    \toprule
    Method & Input Size & Backbone  & Head$\uparrow$  & Sho.$\uparrow$  & Elb$\uparrow$   & Wri.$\uparrow$  & Hip$\uparrow$   & Knee$\uparrow$  & Ank.$\uparrow$  &  PCKh@0.5$\uparrow$  & Params (M)$\downarrow$  & FLOPs (G) $\downarrow$  \\
    \hline

    SimBase~\cite{xiao2018simple} & \multirow{8}[2]{*}{32x32} & ResNet-152 & 40.894 & 48.217 & 33.953 & 21.742 & 44.608 & 31.353 & 22.650 & 36.786 & 68.6  & 0.328 \\
    SimBase~\cite{xiao2018simple} &  & ResNet-50 & 41.473 & 49.474 & 34.839 & 22.393 & 45.837 & 32.541 & 23.405 & 37.780 & 34.0  & 0.187 \\
    HRNet~\cite{sun2019deep} &       & HRNet-W32 & 46.044  & 52.378  & 40.191  & 28.407  & 48.780  & 37.176  & 27.538  & 42.217  & 28.5  & 0.148 \\
    Dark~\cite{zhang2020distribution}  &       & HRNet-W32 & 38.984 & 61.804 & 46.191 & 31.867 & 61.277 & 45.234 & 28.839 & 48.137 & 28.5  & 0.148 \\
    CAL~\cite{wang2022low}  &       & HRNet-W32 & 77.115 & 68.631 & 48.185 & 33.066 & 63.164 & 46.264 & 40.552 & 55.353 & 49.3  & 0.209 \\
    Tokenpose~\cite{li2021tokenpose} &       & HRNet-W32 & 38.950 & 61.702 & 47.537 & 31.902 & 61.676 & 45.476 & 28.600  & 48.421 & 30.9  & 0.152 \\
    
    PRTR~\cite{li2021pose}  &       & HRNet-W32 & 0.341 & 1.512 & 8.062 & 9.082 & 15.977 & 1.633 & 0.213 & 5.618 & 57.2  & 1.476 \\
    \rowcolor{gray!10}Simcc (\textit{baseline})~\cite{li2022simcc}&       & HRNet-W32 & 81.105  & 72.690  & 54.219  & 37.898  & 63.822  & 51.380  & 46.552  & 59.560  & 28.5  & 0.148 \\
    \textbf{CDKD (\textit{ours})}  &       & HRNet-W32 & \textbf{83.049} & \textbf{74.711} & \textbf{56.213} & \textbf{41.427} & \textbf{64.999} & \textbf{52.589} & \textbf{48.724} & \textbf{61.442\textcolor{blue}{$_{
    +1.882}$}} & \textbf{\textcolor{red}{28.5}}  & \textbf{\textcolor{red}{0.148}} \\
    \midrule
    SimBase~\cite{xiao2018simple} & \multirow{9}[1]{*}{64x64} & ResNet-152 & 89.905 & 85.258 & 73.854 & 64.058 & 76.216 & 68.628 & 62.659 & 75.353 & 68.6 & 1.314 \\
    OKDHP~\cite{li2021online} &       & 4-Stack HG & 85.573 & 80.757 & 66.099 & 54.328 & 71.767 & 61.032 & 54.393 & 69.037 & 30.9  & 2.899 \\
    HRNet~\cite{sun2019deep} &       & HRNet-W32 & 90.075 & 85.785 & 72.950 & 63.511 & 75.143 & 67.257 & 61.880 & 74.843 & 28.5  & 0.593 \\
    Dark~\cite{zhang2020distribution}  &       & HRNet-W32 & 89.802 & 87.568 & 75.251 & 64.112 & 78.726 & 69.616 & 63.628 & 76.612 & 28.5  & 0.593 \\
    CAL~\cite{wang2022low}   &       & HRNet-W32 & 92.497 & 88.689 & 75.763 & 65.019 & 80.076 & 70.159 & 65.422 & 77.692 & 49.3  & 0.837 \\
    Tokenpose~\cite{li2021tokenpose} &       & HRNet-W32 & 89.018 & 86.770 & 71.570 & 59.963 & 77.237 & 65.947 & 59.139 & 73.997 & 31.0    & 0.602 \\
    PRTR~\cite{li2021pose}  &       & HRNet-W32 & 89.768 & 83.967 & 68.144 & 53.061 & 76.112 & 62.039 & 53.660 & 70.742 & 57.2  & 1.476 \\
    \rowcolor{gray!10}Simcc (\textit{baseline})~\cite{li2022simcc} &       & HRNet-W32 
    &93.111 & 88.808 & 77.280 & 67.604 & 79.072 & 70.400 & 66.107 & 78.340   & 28.6 & 0.593  \\
    \textbf{CDKD (\textit{ours})}  &       & HRNet-W32 & \textbf{93.622} & \textbf{89.844} & \textbf{78.422} & \textbf{68.137} & \textbf{80.820} & \textbf{71.991} & \textbf{66.438} & \textbf{79.362}\textcolor{blue}{$_{
    +1.022}$} & \textcolor{red}{28.6}& \textbf{\textcolor{red}{0.593}} \\
    \bottomrule
    \end{tabular}%

  \label{tab01}%
\end{table*}%
$\boldsymbol{p}_j^t$ is the j-th probability values after mergers. Then, the teacher model and the student model achieves class alignment, and logit distillation can be performed between them. Based on the above results, we can calculate the logit distillation loss:
\begin{equation}\begin{aligned}L_{logit}(p^t,p^s,\tau)&=\sum_{j=1}^J\tau^2KL(p_j^t,p_j^s)\end{aligned}\label{llogit}\end{equation}
where $\tau$ is the temperature, $\boldsymbol{p}^t$ and  $\boldsymbol{p}^s$ denote the output probability distribution produced by teacher and student,  $KL(\cdot)$ is the Kullback-Leibler Divergence.
With feature distillation loss $\mathit{L_{fea}}$ and logit distillation loss 
$\mathit{L_{logit}}$, we can train the student with the total loss as:
\begin{equation}\begin{aligned}L_{total}=L_{ori}+\alpha L_{fea}+\beta L_{logit}\end{aligned}\label{ltotal}\end{equation}
where $\mathit{L_{ori}}$ is the original task loss for HPE, $\alpha $ and $\beta$ are the hyperparameters to balance the loss.
\subsubsection{ Easy-to-Hard Training Strategy}

In human education, teachers often employ a curriculum strategy of imparting knowledge from simple to challenging, ensuring that students achieve optimal learning outcomes. 
Influenced by this, many researchers adopt the classic curriculum strategy training the models in an easy-to-hard method, which improves the performance of the models.

Inspired by CTKD~\cite{li2023curriculum},
we adopt an easy-to-hard training (ETHT) strategy to increase the difficulty gradually. 
As shown in Fig.~\ref{fig03}, this strategy is implemented by adding a hyperparameter learning module to the student model. In this work, We choose temperature $\tau$ as the hyperparameter to be trained, which can also be extended to other hyperparameters, such as loss coefficient $\alpha$ and $\beta$ (see Table~\ref{tab06}).


Firstly, this module is optimized in the opposite direction of the student's training, intending to maximize the distillation loss between the student and teacher. It can enhance the training performance of the model by producing adversarial effects.
\begin{table*}[htbp]
  \caption{Comparison with SOTA
  methods on the COCO validation dataset. (*) denotes the experimental results obtained by incrementing the splitting factor's value followed by Simcc~\cite{li2022simcc}.
  }
  \centering
  \renewcommand\arraystretch{1.25}
    \setlength{\tabcolsep}{4.0pt}
    \begin{tabular}{ccccccccccc}
    \toprule
    Method & Input size & Backbone & AP$\uparrow$    & AP50$\uparrow$  & AP75$\uparrow$  & AP(M)$\uparrow$ & AP(L)$\uparrow$ & AR$\uparrow$    & Params (M)$\downarrow$ & FLOPs (G)$\downarrow$  \\
    \hline

        HRNet~\cite{sun2019deep} & \multirow{10}[2]{*}{32x32} & HRNet-W32 & 8.2 & 36.9 & 1.3 & 
    9.2 & 6.8 & 15.0  & 28.5  & 0.148 \\
    TokenPose~\cite{li2021tokenpose} &       & HRNet-W32 & 14.0  & 48.2 & 3.4 & 15.2 & 12.5 & 21.7 & 33.1  & 0.156 \\
    CAL~\cite{wang2022low} &       & HRNet-W32 & 26.4 & 61.9 & 18.2 & 27.1 & 25.7 & 33.9 & 49.3  & 0.209 \\
    Rle~\cite{li2021human}&       & HRNet-W32 & 24.4 & -     & -     & -     & -     & -     & 29.1  & 0.593 \\
    Dark~\cite{zhang2020distribution} &       & HRNet-W32 & 12.5 & 45.2 & 2.5 & 13.8 & 11.1 & 20.3 & 28.5  & 0.148 \\
    SimBaese~\cite{xiao2018simple} &       & ResNet-152 & 4.4 & 21.1 & 1.0  & 5.3 & 3.2 & 9.0  & 68.6  & 0.328 \\
    PCT~\cite{geng2023human} &       & Swin-Base & 1.3 & 4.6 & 10.0   & 1.0 & 1.2 & 3.1 & 86.9  & 1.676 \\
    Distillpose~\cite{ye2023distilpose} &       & HRNet-W32 & 9.5 & 32.6 & 2.4 & 10.0   & 9.5 & 21.2 & 33.0    & 0.310\\
    \rowcolor{gray!10}SimCC (\textit{baseline})~\cite{li2022simcc}&       & HRNet-W32 & 29.8 & 65.6 & 22.5 & 30.0   & 29.9 & 36.3 & 28.5  & 0.148 \\
    \textbf{CDKD (\textit{ours})} &       & HRNet-W32 & \textbf{30.7}\textcolor{blue}{$_{
    +0.9}$} & \textbf{66.4} & \textbf{23.5} & \textbf{30.9} & \textbf{30.7} & \textbf{37.3} & \textcolor{red}{28.5}  & \textcolor{red}{0.148} \\
    \midrule
    SimpleBaseline~\cite{xiao2018simple} & \multirow{9}[2]{*}{64x64} & ResNet-152 & 30.3 &	67.6&	22.6&	30.6 &	30.5 &	36.2& 68.6
    & 1.314 \\
    Distillpose~\cite{ye2023distilpose} &       & HRNet-W32 & 31.7 & 66.8 & 26.6 & 32.3 & 31.8 & 44.5 & 33.1  & 0.790 \\
    PCT~\cite{geng2023human} &       & Swin-Base  & 13.8 & 41.1 & 6.4 & 14.1 & 14.2 & 19.6 & 86.9  & 2.300 \\
    Rle~\cite{li2021human}   &       & HRNet-W32 & 52.5 & -     & -     & -     & -     & -     & 29.1  & 0.593 \\
    HRNet~\cite{sun2019deep} &       & HRNet-W48 & 46.9 & 83.7 & 49.2 & 46.6 & 47.5 & 52.6 & 63.6  & 1.218 \\
    Tokenpose~\cite{li2021tokenpose} &       & HRNet-W48 & 50.3 & 82.7 & 54.4 & 49.9 & 51.4 & 55.6 & 68.2  & 1.236 \\
    CAL~\cite{wang2022low}   &       & HRNet-W48 & 60.6 & \textbf{88.1} & 68.4 &59.5 &62.3 & 65.5 & 110.3 & 1.797\\
    Dark~\cite{zhang2020distribution}  &       & HRNet-W48 & 57.2 & 86.8 & 63.5 & 55.9 & 59.2 & 62.2 & 63.6  & 1.218 \\
    \rowcolor{gray!10}Simcc (\textit{baseline})~\cite{li2022simcc} &       & HRNet-W48 & 58.6 & 85.9 & 64.9 & 57.8 & 60.5 & 63.4 & 63.7  & 1.218 \\
    \textbf{CDKD (\textit{ours})} &       & HRNet-W48 & 60.3\textcolor{blue}{$_{
    +1.7}$} & 86.9 & 67.3 & 59.6 & 62.0  & 65.0  & 63.7  & 1.218 \\
    \rowcolor{gray!10}Simcc (\textit{baseline})*~\cite{li2022simcc} &       & HRNet-W48 & 59.7 & 85.0 & 67.3 & 58.4 & \textbf{64.0} & \textbf{67.5} & 63.7  & 1.218 \\
    \textbf{CDKD (\textit{ours})}* &       & HRNet-W48 & \textbf{61.1}\textcolor{blue}{$_{
    +1.4}$} & 86.9 & \textbf{68.4} & \textbf{59.9} & 62.9  & 65.6  & \textcolor{red}{63.7}  & \textcolor{red}{1.218} \\
    \bottomrule
    \end{tabular}%

  \label{tab02}%
\end{table*}%
We apply the gradient reversal method to implement the adversarial process. The traditional gradient descent process is as follows: 
\begin{equation}\begin{aligned}\theta&\leftarrow\theta-\eta\frac{\partial L}{\partial\theta}\end{aligned}\end{equation}
where $\theta$ is the training parameters, $\eta$ is the is the learning rate, and $\mathit{L}$ is the loss. Our proposed gradient reversal method is as follows:
\begin{equation}\begin{aligned}\theta_{hype}&\leftarrow\theta_{hype}+\eta\frac{\partial L}{\partial\theta_{hype}}\end{aligned}\end{equation}
where $\theta_{hype}$ is the training hyperparameters. We optimize the hyperparameter module in the opposite direction of gradient descent, which essentially prevents loss from declining and increases the difficulty of model training. In addition, to gradually increase the training difficulty, we introduce a
dynamic coefficient $\xi$ in the above process of backpropagation.
\begin{equation}\begin{aligned}\theta_{hype}&\leftarrow\theta_{hype}+\eta\frac{\partial (\xi L)}{\partial\theta_{hype}}\end{aligned}\end{equation}
\begin{equation}\xi=\mathrm{\mathit{func}}(\frac{T_i}{T_{\max}})\end{equation}
where $\mathit{func(\cdot)}$ denotes the monotonically increasing function, $\mathit{T_i}$ is current epoch, and $\mathit{T_{max}}$ is total epochs.
As training progresses, $\xi$ gradually increases, and the training difficulty gradually increases. Therefore, by adopting this approach, the training performance is effectively optimized.

\section{Experiments}
In this section, we first evaluate the proposed distillation
framework on two common benchmark datasets:  MPII~\cite{andriluka20142d} and COCO~\cite{lin2014microsoft}. Then we
carry out a series of ablation studies to prove the effectiveness and validity of our approach.
\begin{table}[t]
    \caption{Ablation studies for different modules on MPII validation set. The ``NECM" means the non-ETHT CCA module. All experiments use HRNet-W32 as the backbone and take images with 64$\times$64 resolution as input.}
  \centering
    \setlength{\tabcolsep}{1.0pt}
    \renewcommand{\arraystretch}{1.2}
    \begin{tabular}{ccccccc}
    \toprule
    & \multicolumn{3}{c}{SAPE}   & \multicolumn{2}{c}{CCA}    & \\ \cmidrule(r){2-4} \cmidrule(r){5-6}
     \multirow{-2}*{Method} & Projector &  Ensemble   & SAU  & NECM    & ETHT  & \multirow{-2}*{PCKh@0.5$\uparrow$} 
    \\
    \midrule
    \rowcolor{gray!10}Simcc \textit{(baseline)}~\cite{li2022simcc} &   -    &   -    &   -    &   -    &   -    & 78.340 \\
    ours  & \checkmark     &       &       &       &       & 78.621 \\
    ours  & \checkmark     & \checkmark     &       &       &       & 78.717 \\
    ours  & \checkmark     & \checkmark     & \checkmark     &       &       & 78.808 \\
    \midrule
    ours  &       &       &       & \checkmark     &       & 78.772
    \\
    ours  &       &       &       & \checkmark    & \checkmark     & 78.811
    \\
    \midrule
    ours  & \checkmark     & \checkmark     & \checkmark    & \checkmark     & \checkmark    & \textbf{79.362}
    \\
    \bottomrule
    \end{tabular}%

  \label{tab03}%
\end{table}%
In addition, we show the universality of our proposed framework in various human pose estimation models.
\subsection{Implementation Details}
\subsubsection{Datasets}
The MPII dataset includes approximately 25K
images containing over 40K subjects with annotated body
joints, where 29K subjects are used for training and 11K
subjects are used for testing. We adopt the same train/valid/test
split as in~\cite{zhang2019fast}. Each person instance in MPII has 16 labeled
joints. The PCKh@0.5 is used for the MPII dataset,
which refers to a threshold of 50\% of the head diameter.

 The COCO dataset contains over 200k images and 250k human instances. Each human instance is
labeled with K = 17 keypoints representing a human pose.
Our models are trained on COCO train2017 with 57k
images and evaluated on COCO val2017, which contain 5k and 20k images, respectively.
We mainly report the commonly
used standard evaluation metrics Average Precision (AP) and Average Recall (AR), which are calculated based on Object Keypoint Similarity (OKS) on the COCO dataset.

\subsubsection{Training Details}
We adopt the two-stage top-down human pose estimation pipeline.  Firstly, the person instances are detected, and then the keypoints are estimated. In the stage of keypoint estimation, we set the cropped human images to low resolution. \textbf{\textit{We replicate SOTA human pose estimation methods under low-resolution conditions, following the settings detailed in their respective papers and code. When reproducing these SOTA researches, we only modify the resolution-related settings, leaving the others unchanged.}} For all methods, we train and test under the same low
resolution. 
We adopt a commonly used person detector provided by SimpleBaselines~\cite{xiao2018simple} with 56.4\% AP for the MSCOCO val dataset.
We utilize Simcc~\cite{li2022simcc} as the base model for the teacher and the student, which uses HRNet~\cite{sun2019deep} as its backbone.
In the case of SimpleBaseline~\cite{xiao2018simple}, the base learning rate is initialized as 1$\mathit{e}$-3. 
It is then decreased to 1$\mathit{e}$-4 and 1$\mathit{e}$-5 at the 90-th and 120-th epochs respectively. 
\begin{table}[htbp]
    \caption{Ablation studies for different modules on COCO val set. The ``NECM" means the non-ETHT CCA module. All experiments use HRNet-W32 as the backbone and take images with 32$\times$32 resolution as input. }
  \centering
    \setlength{\tabcolsep}{1.5pt}
    \renewcommand{\arraystretch}{1.2}
    \begin{tabular}{ccccccc}
    \toprule
    & \multicolumn{3}{c}{SAPE}   & \multicolumn{2}{c}{CCA}    & \\ \cmidrule(r){2-4} \cmidrule(r){5-6}
     \multirow{-2}*{Method} & Projector &  Ensemble   & SAU  & NECM    & ETHT  & \multirow{-2}*{AP$\uparrow$}  
    \\ \hline
    \rowcolor{gray!10}Simcc \textit{(baseline)}~\cite{li2022simcc} &   -    &   -    &   -    &   -    &   -    & 
    29.8 \\
    ours  & \checkmark     &       &       &       &       & 30.3 \\
    ours  & \checkmark     & \checkmark     &       &       &       & 30.4 \\
    ours  & \checkmark     & \checkmark     & \checkmark     &       &       & 30.6 \\
    \midrule
    ours  &       &       &       & \checkmark     &       & 30.3
    \\
    ours  &       &       &       & \checkmark    & \checkmark     & 30.6
    \\
    \midrule
    ours  & \checkmark     & \checkmark     & \checkmark    & \checkmark     & \checkmark    & \textbf{30.7}
    \\
    \bottomrule
    \end{tabular}%
  \label{tababcoco}%
\end{table}%

\begin{table}[htbp]
  \caption{Ablation studies of the alignment method
on the MPII validation dataset.
``Conv Downsample" and ``Fully Connected Layer" mean conventional feature and class alignment methods, respectively.
``Same Class Number" refers to the method of making the number of categories consistent between the teacher model and the student model by altering the splitting factor~\cite{li2022simcc}. }
  \centering
   \setlength{\tabcolsep}{6pt}
    \renewcommand{\arraystretch}{1.2}
    \begin{tabular}{ccc}
    \toprule
    Distillation & Method & PCKh@0.5$\uparrow$ \\
    \hline
    \rowcolor{gray!10}\multirow{4}[2]{*}{Feature distillation} 
    & Conv Downsample~\cite{romero2014fitnets} & 78.207 \\
          & Fully Connected Layer~\cite{zhu2023scalekd} & 78.249 \\
          & Projector Ensemble 
          \textit{(ours)} & 78.717 \\
          & SAPE \textit{(ours)} & \textbf{78.808} \\
    \midrule
    \rowcolor{gray!10}\multirow{4}[2]{*}{Logit distillation} & Same Class Number~\cite{li2022simcc} & 75.173 \\
          & Fully Connected Layer~\cite{ge2020efficient}   & 77.973 \\
          & NECM \textit{(ours)} & 78.772 \\
          & CCA \textit{(ours)} & \textbf{78.811} \\
    \bottomrule
    \end{tabular}%
  \label{tab04}%
\end{table}%
As for HRNet~\cite{sun2019deep}, the base learning rate is also initialized as 1$\mathit{e}$-3. It is subsequently reduced to 1$\mathit{e}$-4 and 1$\mathit{e}$-5 at the 170-th and 200-th epochs respectively. The total training processes conclude at the 140th and 210th epochs respectively for SimpleBaseline~\cite{xiao2018simple} and HRNet~\cite{sun2019deep}.
The training settings for CDKD, such as the optimizer,
learning rate, and data augmentation, are the same as Simcc~\cite{li2022simcc}. Experiments are conducted on 2 NVIDIA 3080-Ti GPUs.

\subsection{Main Results}

\subsubsection{Results on MPII Dataset}
We evaluate the CDKD framework on the MPII validation dataset. Table~\ref{tab01}
compares the PCKh@0.5 accuracy results of CDKD and the SOTA methods under low-res conditions. We can clearly observe that Simcc~\cite{li2022simcc} achieves outstanding performance with low computational cost. Adding CDKD, Simcc~\cite{li2022simcc} is further improved.
It achieves better accuracy than any 
other method. Specifically, in resolutions of 64x64 and 32x32, our method improves the baseline network by 1.022\% and 1.882\%, respectively. 
\begin{table}[htbp]
  \caption{Ablation studies of the projector number on the MPII validation dataset. The projector number refers to the number of projectors in the SAPE module.
  }
  \centering
  \setlength{\tabcolsep}{2pt}
    \renewcommand{\arraystretch}{1.2}
    \begin{tabular}{cccccc}
    \toprule
    Projector number & 1     & 2     & 3     & 4     & 5 \\
    \midrule
    PCKh@0.5$\uparrow$ & 78.621 & 79.037 & \textbf{79.362} & 76.937 & 76.831 \\
    \bottomrule
    \end{tabular}%
  \label{tab07}%
\end{table}%

\begin{table}[htbp]
  \caption{Ablation studies of the number of convolutions with different kernel sizes in SAU on the MPII validation dataset. The 3x3, 5x5, 7x7, and 9x9 refer to the conv kernel sizes. }
  \centering
    \setlength{\tabcolsep}{8pt}
    \renewcommand{\arraystretch}{1.2}
    \begin{tabular}{cccccc}
    \toprule
    Conv & 3x3   & 5x5   & 7x7   & 9x9   & PCKh@0.5$\uparrow$ \\
    \midrule
       \rowcolor{gray!10}& 1     & 1     & 1     & 0     & 78.972 \\
    Conv number  & 2  & 1     & 1     & 0     & 78.884 \\
      & 3  & 1     & 1     & 0     & 78.519 \\
    \midrule
     & 1     & 2  & 1     & 0     & 79.173  \\
    Conv number  & 1     & 1     & 2  & 0     & \textbf{79.362}  \\
     & 1     & 1     & 1     & 1  & 79.027 \\
    \bottomrule
    \end{tabular}%

    \label{tab05}%
\end{table}%
 It does not incur any additional costs in parameters and GFLOPs,
The performance is significant as compared to prior works. In short, CDKD  achieves SOTA performance with \textit{\textbf{NO}} increase in computational cost.
\subsubsection{Results on COCO Dataset}
Table~\ref{tab02} shows the results of the cutting-edge methods and CDKD under low-res conditions on the COCO val2017 set. In resolutions of 32$\times$32 and 64$\times$64, our method improves the baseline model 0.9\% and 1.7\%, respectively. Especially, it achieves SOTA  performance, whether in a resolution of 32$\times$32 or 64$\times$64.
Meanwhile, it doesn't bring any increase in computation and parameters. Our proposed CDKD method is an efficient and effective approach in low-res HPE.


\subsection{Ablation Study}

\textbf{Different modules.} In this subsection, we conduct several ablation experiments
to show how each module helps the
training of the low-res student. As shown in Table~\ref{tab03}, all modules benefit the
low-res model. SAPE and CCA bring an
improvement of 0.468\% and 0.471\%, respectively. 
In SAPE,
%
 the scale-adaptive unit (SAU) further improves the performance of the projector ensemble by 0.091\%.
Meanwhile, the CCA module brings
the student 0.432\% gains. When combing the ETHT strategy, the gains get to 0.471\%.
The combination of
all proposed modules bring the best performance, which significantly improves the performance by 1.022\%.
\begin{table}[t]
    \caption{Ablation studies of the hyperparameters in the ETHT strategy on the MPII validation dataset. The $\tau$, $\alpha$, and $\beta$ refer to key hyperparameters in the CDKD framework. }
  \centering
   \setlength{\tabcolsep}{8pt}
    \renewcommand{\arraystretch}{1.2}
    \begin{tabular}{ccc}
    \toprule
    Method & Hyperparameters & PCKh@0.5$\uparrow$ \\
    \midrule
     \rowcolor{gray!10}Non-ETHT
     &   -  &    78.772\\
    ETHT  &$\tau$     & \textbf{79.362} \\
    ETHT  & $\tau,~\alpha$   & 79.053 \\
    ETHT  & $\tau$,~$\alpha$,~$\beta$ & 79.206 \\
    \bottomrule
    \end{tabular}%

  \label{tab06}%
\end{table}%

In order to thoroughly demonstrate the performance of the various modules proposed by us, we also conduct ablation experiments on the COCO dataset. As shown in Table~\ref{tababcoco}, each module improves the model's performance. The combination of all modules achieves the optimal performance for the model.


\begin{figure*}[!t]
\centering
 \captionsetup[subfigure]{labelformat=empty}
 \subfloat[]{\includegraphics[width=0.16\linewidth,height=1.3in]{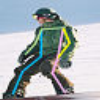} }
 \subfloat[]{\includegraphics[width=0.16\linewidth,height=1.3in]{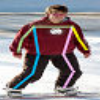} }
 \subfloat[]{\includegraphics[width=0.16\linewidth,height=1.3in]{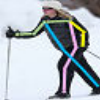} }
 \subfloat[]{\includegraphics[width=0.16\linewidth,height=1.3in]{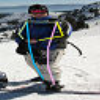} }
  \subfloat[]{\includegraphics[width=0.16\linewidth,height=1.3in]{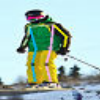} }
   \subfloat[]{\includegraphics[width=0.16\linewidth,height=1.3in]{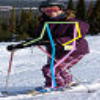} }
     \vspace{-0.3in}
    \subfloat[]{\includegraphics[width=0.16\linewidth,height=1.3in]{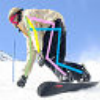} }
 \subfloat[]{\includegraphics[width=0.16\linewidth,height=1.3in]{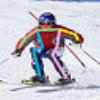} }
 \subfloat[]{\includegraphics[width=0.16\linewidth,height=1.3in]{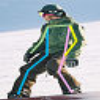} }
 \subfloat[]{\includegraphics[width=0.16\linewidth,height=1.3in]{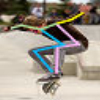} }
  \subfloat[]{\includegraphics[width=0.16\linewidth,height=1.3in]{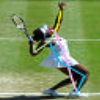} }
   \subfloat[]{\includegraphics[width=0.16\linewidth,height=1.3in]{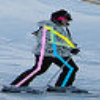} }
 


	\caption{Visual results of our CDKD framework on the COCO val dataset. We adopt Simcc~\cite{li2022simcc} as the base model. The resolution of all images is 64$\times$64. }
	\label{figvis_2}
\end{figure*}


\textbf{Alignment methods.} 
We compare the results of CDKD with conventional feature alignment and class alignment in Table~\ref{tab04}. Projector ensemble and NECM both surpass conventional methods. On this basis, SAPE and CCA further improve the model's performance. It indicates our approach significantly outperforms traditional alignment methods.

\textbf{Projector number.} We evaluate the impact of projector number on
the performance of CDKD in Table~\ref{tab07}. 
The use of a projector ensemble improves the model's performance, but an excessive number of projectors would result in a decline in performance.
As shown in Table~\ref{tab07}, the CDKD reaches the best performance when the number of projectors in SAU is 3.

\textbf{Conv number.} 
We compare the impact of the convolution number of different kernel sizes in our proposed SAU module, as shown in Table~\ref{tab05}. we can observe that increasing moderately
the proportion of convolutions with a large receptive field can further improve the performance of our proposed CDKD distillation framework.
It fully demonstrates that our designed module adequately captures multi-scale human body information.

\textbf{Hyperparameters.} Table~\ref{tab06} shows how the performance of our proposed framework is affected
by the choice of hyperparameters in Eq.~\ref{llogit} and Eq.~\ref{ltotal}. 
It indicates that the ETHT strategy we designed is effective.
Each choice of hyperparameters helps to improve the model performance.
When selecting $\tau$ as a learning object,
our method reaches the best result.

\begin{table}
  \caption{Evaluation of the CCA module in our proposed CDKD framework applied under different conditions on the COCO val dataset.}
  \centering
    \setlength{\tabcolsep}{2pt}
    \renewcommand{\arraystretch}{1.2}
    \begin{tabular}{cccccc}
    \toprule
    Methods & Resolution & Role &Backbone & AP$\uparrow$    & AR$\uparrow$ \\
    \midrule
    Simcc~\cite{li2022simcc} &256$\times$256&teacher& HRNet-48 & 76.4    & 79.4\\
    \rowcolor{gray!10}Simcc \textit{(baseline)}~\cite{li2022simcc} &64$\times$64&-& ResNet-50 & 40.0    & 45.5 \\
    CCA  &64$\times$64&student& ResNet-50 & \textbf{42.0\textcolor{red}{$_{
    +2.0}$}} & \textbf{47.6\textcolor{red}{$_{+2.1}$}}\\

    \bottomrule
    \end{tabular}%

  \label{tabcca}
\end{table}%

\begin{table}
   \caption{
   Evaluation of the SAPE module in our proposed CDKD framework applied in different models on MPII val
dataset. For CAL, the teacher model is trained using the CAL method with a resolution of 256$\times$256. For HRNet, the teacher model is trained using the HRNet method with a resolution of 256$\times$256.
   }
  \centering
    \setlength{\tabcolsep}{8pt}
    \renewcommand{\arraystretch}{1.2}
    \begin{tabular}{cccc}
    \toprule
    Methods & Backbone & Resolution & PCKh@0.5$\uparrow$  \\
    \rowcolor{gray!10}CAL~\cite{wang2022low}   & HRNet  & 32$\times$32 & 55.353 \\
    CAL (+SAPE) &   -    & 32$\times$32 & \textbf{56.029\textcolor{red}{$_{
    +0.676}$}}\\
    
    \rowcolor{gray!10}HRNet~\cite{sun2019deep} &   -    & 64$\times$64 & 74.843 \\
    HRNet (+SAPE) &   -    & 64$\times$64 & \textbf{76.248\textcolor{red}{$_{
    +1.405}$}}\\
    \bottomrule
    \end{tabular}%
  \label{tabsape}%
\end{table}%

\subsection{Universality} 
We further extend our proposed CDKD framework to other pose estimation models to illustrate its universality. 
In current research on human pose estimation, 
the scope of the CCA module and the SAPE module differs significantly.
Compared to the CCA module, SAPE is applicable to almost all models.
Therefore, we conduct separate research on the CCA module and the SAPE module.

\textbf{CCA module.}  We extend the CCA module to different models. As shown in Table~\ref{tabcca}, the teacher model and the student model employ distinct backbones and have varying numbers of classes.
The teacher model adopts HRNet-48 as the backbone, while the student model adopts ResNet-50. 
The two models have noticeable differences.
In this case, the student model is still well-optimized by the CCA module, demonstrating the universality of our CCA module.

\textbf{SAPE module.}  As an easy-to-use
plug-in technique, SAPE can be seamlessly integrated into
existing HPE works. As shown in Table~\ref{tabsape}, our method
brings comprehensive improvements to two SOTA
methods based on two teacher-student pairs. It fully demonstrates the generality and effectiveness of our approach.

\subsection{Visualized Results}
Fig.~\ref{figvis_2} provides visualized human pose estimation results. As is shown in Fig.~\ref{figvis_2}, the CDKD model achieves accurate human pose estimation when the input is low-resolution.

\subsection{Limitation and Future Work}
As a novel universal distillation framework, the proposed CDKD opens up many possible directions for future works to address the bottleneck limitations.
Our SAPE module can be extended to various knowledge distillation tasks, enabling the distillation between features with different sizes.
The CCA module we propose addresses the issue of inconsistent categories.
It opens up a new perspective on distillation across categories. 
Nonetheless, our cross-domain knowledge distillation method is presently confined to distilling between conventional models. In the future, we plan to apply our method to large models, thus maximizing the utilization of their abundant knowledge.

\section{Conclusion}
In this work, we propose a novel low-resolution human pose estimation framework (CDKD), which includes a  scale-adaptive projector ensemble (SAPE) module and a cross-class alignment
(CCA) module to perform high-resolution to low-resolution knowledge distillation. In this
way, the student model acquires richer image
knowledge at both feature and logit levels, achieving a big
leap in performance while maintaining efficiency. Extensive experiments conducted on the COCO and MPII datasets
demonstrate the effectiveness of our method.
In short, CDKD achieves SOTA performance
among low-res methods with a low computational cost.

\bibliographystyle{IEEEtran}
\bibliography{IEEEabrv,mylib}











\newpage

 




\vfill

\end{document}